\documentclass[preprint,12pt]{elsarticle}
\usepackage{color}
\usepackage{graphicx}
\usepackage{verbatim}
\usepackage{amsmath,amssymb}
\RequirePackage{lineno}
\usepackage{fancyhdr}
\usepackage{subfigure}
\usepackage{lscape}
\usepackage{verbatim}
\usepackage{amssymb}
\usepackage{color}
\usepackage[ruled]{algorithm2e}

\begin{document}
\title{\textrm{Interpreting Tree Ensembles with inTrees}}
\author{Houtao Deng}
\ead{softwaredeng@gmail.com}
\address{California, USA}

\begin{abstract}
Tree ensembles such as random forests and boosted trees are accurate but difficult to understand, debug and deploy.
In this work, we provide the inTrees (interpretable trees) framework that extracts, measures, prunes and selects rules from a tree ensemble, and calculates frequent variable interactions. An rule-based learner, referred to as the simplified tree ensemble learner (STEL), can also be formed and used for future prediction. The inTrees framework can applied to both classification and regression problems, and is applicable to many types of tree ensembles, e.g., random forests, regularized random forests, and boosted trees. We implemented the inTrees algorithms in the ``inTrees" R package.
\end{abstract}


\maketitle

\newtheorem{Definition}{Definition}
\newtheorem{Lemma}{Lemma}

\section{Introduction}
Let $X=(X_1,...,X_p)$ be the predictor variables, $T$ be the target (or outcome). The goal of supervised learning is to build a model using $X$ to predict $T$. It is called a classification problem when $T$ is discrete and a regression problem when $T$ is numeric.

Tree ensembles such as random forests \cite{breiman2001random} and boosted trees \cite{friedman2001greedy} are accurate supervised learners, and so are powerful in capturing information in the data. However, They are difficult to a) \textbf{understand}; It may be hard, particularly for people without modeling background, to rely on a model without understanding it. b) \textbf{debug}; Just like software engineers can make bugs in programming, data analysts can make ``modeling bugs" (also referred to as ``leakage" in \cite{kaufman2012leakage}), i.e., building a model in an inappropriate way so the model does not function as expected. It is hard to discover bugs in a model without interpretable information from the model. c) \textbf{deploy}. It is not easy to code a tree ensemble with potentially a huge number of trees, particularly for situations where models are trained off-line in one language such as R, but are applied on-line using another language such as Java.

In this work we propose to extract interpretable information from tree ensembles, referred to as the \emph{inTrees} (interpretable trees) framework. Particularly, we propose methods to extract, measure and process rules from a tree ensemble, and extract frequent variable interactions. Also, the rules from a tree ensemble are used to build a rule-based learner for future predictions.

\def\myWidth{3.5}
\begin{figure}[!]
\centering
    \includegraphics[width= \myWidth in]{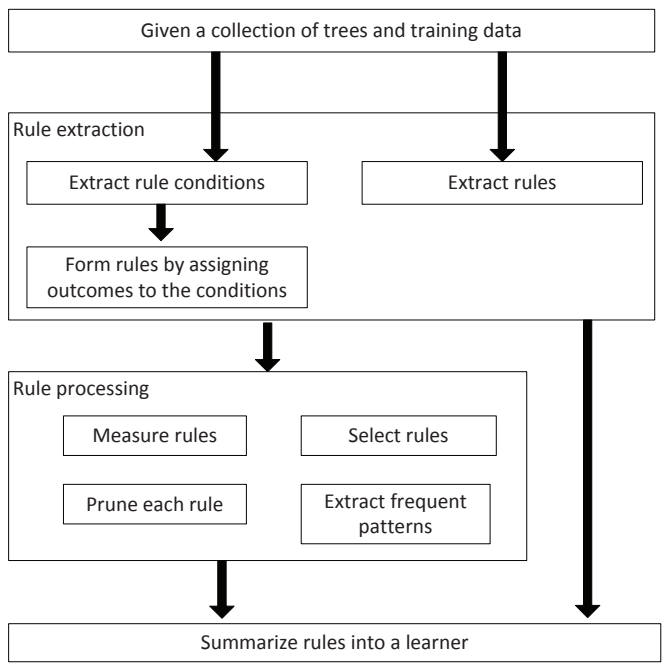}
\caption{Illustration of the inTrees framework.\label{fig:intreesframework}}
\end{figure}

\section{The inTrees Framework}
We propose the inTrees (interpretable trees) framework to extract insights from tree ensembles. Particularly, inTrees consists of algorithms to extract rules, measure (and thus rank) rules, prune irrelevant or redundant variable-value pairs of a rule (prune each rule), select a compact set of relevant and non-redundant rules (select rules), discover frequent variable interactions (extract frequent patterns), and summarize rules into a learner that can be used for predicting new data. The functions in the framework and the relationship among the functions are shown in Figure \ref{fig:intreesframework}. One can extract rules from a tree ensemble, perform rule processing such as pruning each individual rule, and then summarize the rules into a learner. One can also directly summarize the rules after rule extraction, or just perform rule extraction and rule processing without the rule summarizing step. The inTrees framework can be applied to both classification and regression problems. Furthermore, the framework can be applied to tree ensembles with decision trees that split each internal node using one feature, and assign the outcome (or target) at leaf nodes. Many popularly used tree ensembles belong to this type, e.g., random forests \cite{breiman2001random}, regularized random forests \cite{deng2013gene}, and boosted trees \cite{friedman2001greedy}.

The inTrees framework is independent from the tree ensemble building process. Therefore, as long as each tree in a tree ensemble is transformed to a specific format, inTrees algorithms can be applied. In the inTrees R implementation, the tree format is defined as the same as the ``randomForest"\cite{liaw2002classification} and ``RRF"\cite{deng2013gene} R packages shown in Table \ref{tab:treePresentation}. The first column in the table is the index of the current node. The next two columns are the left and right daughter nodes of the current node, respectively. Column 4 - 5, respectively, are the split variable and split value at the current node. The ``status" column is ``-1" for a leaf node, and other values for non-leaf nodes. The ``pred" column is the prediction at a leaf node, that is, the assignment of the outcome. Note although the inTrees algorithms can be applied to a node with more than two children nodes, the inTrees R package (version 1.0) currently only handles binary splits.

Also, inTrees algorithms can be applied to each tree in a tree ensembles in parallel, thus can be implemented in distributed computing environments.

The methods in the inTrees framework are introduced in the following sections.

\begin{table}[h]
\begin{center}
\scriptsize
\begin{tabular}{|c|c|c|c|c|c|c|}
\hline
      node & left  & right  &  split & split      & status & pred \\
           &  daughter & daughter &  var &  point &   &  \\
\hline
         1 &          2 &          3 &          5 &          2 &          1 &          0 \\
\hline
         2 &          4 &          5 &          1 &          3 &          1 &          0 \\
\hline
         3 &          6 &          7 &          3 &          2 &          1 &          0 \\
\hline
         4 &          8 &          9 &          9 &          3 &          1 &          0 \\
\hline
         5 &         10 &         11 &          2 &          2 &          1 &          0 \\
\hline
         6 &         12 &         13 &          1 &          2 &          1 &          0 \\
\hline
         7 &         14 &         15 &          7 &          2 &          1 &          0 \\
\hline
         8 &          0 &          0 &          0 &          0 &         -1 &          1 \\
\hline
         9 &         16 &         17 &          4 &          1 &          1 &          0 \\
\hline
        10 &         18 &         19 &          3 &          3 &          1 &          0 \\
\hline
        11 &         20 &         21 &          3 &          3 &          1 &          0 \\
\hline
        12 &         22 &         23 &          2 &          2 &          1 &          0 \\
\hline
        13 &         24 &         25 &          9 &          2 &          1 &          0 \\
\hline
        14 &         26 &         27 &          9 &          2 &          1 &          0 \\
\hline
        15 &          0 &          0 &          0 &          0 &         -1 &          2 \\
\hline
        16 &         28 &         29 &          7 &          2 &          1 &          0 \\
\hline
        17 &         30 &         31 &          6 &          2 &          1 &          0 \\
\hline
        18 &         32 &         33 &          7 &          3 &          1 &          0 \\
\hline
        19 &         34 &         35 &          8 &          1 &          1 &          0 \\
\hline
\end{tabular}
\caption{General tree presentation in the inTrees R package (version 1.0).\label{tab:treePresentation}}
 \end{center}
 \end{table}

\IncMargin{1em}
\begin{algorithm}
\scriptsize
\LinesNumbered
\SetKwData{Left}{left}\SetKwData{This}{this}\SetKwData{Up}{up}
\SetKwFunction{Union}{Union}\SetKwFunction{FindCompress}{FindCompress}
\SetKwInOut{Input}{input}\SetKwInOut{Output}{output}
\Input{$ruleSet \leftarrow null, node \leftarrow rootNode, C \leftarrow null$}
\Output{$ruleSet$}
\BlankLine
\If{$leafNode = true$}{\label{lt}
$currentRule \leftarrow \{C \Rightarrow  pred_{node}$\} \\
$ruleSet \leftarrow \{ ruleSet, currentRule\}$ \\
\Return{ruleSet}
}
\For{$child_i$ = every child of $node$}{
 $C \leftarrow C \wedge C_{node}$ \\
 $ruleSet \leftarrow ruleExtract(ruleSet,child_i,C)$
}
\Return{ruleSet}
\caption{\scriptsize{$ruleExtract(ruleSet,node,C)$: function to extract rules $ruleSet$ from a decision tree. In the algorithm, let $C$ denote the conjunction of variable-value pairs aggregated from the path from the root node to the current node, $C_{node}$ denote the variable-value pair used to split the current node, $leafNode$ denote the flag whether the current
node is a leaf node, and $pred_{node}$ denote the prediction at a leaf node.} }\label{algo:extractrule}
\end{algorithm}\DecMargin{1em}

\section{Extract Rules}\label{sec:extract}
A tree ensemble consists of multiple decision trees \cite{breiman2001random,friedman2001greedy}. 
A rule can be extracted from a decision tree's root node to a leaf node. A rule can be expressed as $\{C \Rightarrow T\}$, where $C$, referred to as the condition of the rule, is a conjunction of variable-value pairs, and $T$ is the outcome of the rule.  Algorithm \ref{algo:extractrule} shows one way to extract rules from a decision tree. The rules extracted from a tree ensemble are a combination of rules extracted each decision tree in the tree ensemble.

Since the outcome values of the rules from a tree ensemble are often assigned based on a part of the training data (e.g., random forest), the assignment may not be reliable. Thus, one can extract the conditions from a tree ensemble, and then assign the outcome values to the conditions based on all the training data. Algorithm \ref{algo:extractcond} shows one way to extract conditions from a decision tree. Note, the most informative splits often occur in the top level of a tree, so one can stop the extraction process when a certain depth of the tree ($maxDepth$ in the algorithm) is reached, which is also more computationally efficient. 
To determine the outcome value of a condition, one can choose the most frequent class for a classification problem, and the mean (or other metrics such as median) for a regression problems, of the training instances that satisfy the condition.

Extracting only conditions from a tree ensemble (i.e., omitting the outcome values assigned by the tree ensemble) also has the following benefit. Say we have built a tree ensemble for regression. If one wants to derive more descriptive rules such as \{$X_1=1 \Rightarrow large$\}, instead of using a continuous value as the outcome, one can simply discretize the target, and then assign discretized levels to the conditions (same as assigning the outcome to a condition for classification problems).

\IncMargin{1em}
\begin{algorithm}
\scriptsize
\LinesNumbered
\SetKwData{Left}{left}\SetKwData{This}{this}\SetKwData{Up}{up}
\SetKwFunction{Union}{Union}\SetKwFunction{FindCompress}{FindCompress}
\SetKwInOut{Input}{input}\SetKwInOut{Output}{output}
\Input{$condSet \leftarrow null, node \leftarrow rootNode, C \leftarrow null$,
$maxDepth \leftarrow -1, curretDepth \leftarrow 0$}
\Output{$condSet$}
\BlankLine
$curretDepth = curretDepth + 1$ \\
\If{$leafNode = true$ or $curretDepth = maxDepth$ }{\label{lt}
$condSet \leftarrow \{ condSet, currentCond\}$\\
\Return{condSet}
}
\For{$child_i$ = every child of $node$}{
 $C \leftarrow C \wedge C_{node}$ \\
 $condSet \leftarrow condExtract(condSet,child_i,C,maxDepth,curretDepth)$
}
\Return{condSet}
\caption{\scriptsize{$condExtract(condSet,node,C,maxDepth,curretDepth)$: function to extract conditions $condSet$ from a tree ensemble. In the algorithm, let $C$ denote the conjunction of variable-value pairs aggregated in the path from the root node to the current node, $C_{node}$ denote the variable-value pair used to split the current node, $leafNode$ denote the flag whether the current
node is a leaf node. Note one can set a maximum depth $maxDepth$ of the tree where the conditions are extracted from. In a decision tree, most useful splits tend to happen in top levels of the trees (i.e., when $depth$ is small), so setting a maximum depth can reduce computations, and may also avoid extracting overfitting rules. $maxDepth$=-1 means there is no limitation on the depth.}}\label{algo:extractcond}
\end{algorithm}\DecMargin{1em}

\section{Measure rules}
Here we introduce metrics to measure rules' quality.  \textbf{Frequency} of a rule is defined as the proportion of data instances satisfying the rule condition. The frequency measures the popularity of the rule. \textbf{Error} of a rule is defined as the number of incorrectly classified instances determined by the rule divided by the number of instances satisfying the rule condition for classification problems, and mean squared error for regression problems,

 \begin{equation}
MSE = \frac{1}{k} \sum_{i=1}^k (t_i-\bar{t})^2
\end{equation}

where $k$ is the number of instances that satisfy the rule condition, and $t_i$ ($i$=1,...,$k$) are the target values of the instances satisfying the condition, and $\bar{t} = \frac{1}{k} \sum_{i=1}^k (t_i)$.

Complexity of a rule is measured by the \textbf{length} of the rule condition, defined as the number of variable-value pairs in the condition. Given two rules with similar frequency and error, the rule with a smaller length may be preferred as it is more interpretable.

By using these definitions, one can rank the rules according to length, support, error or a combination of multiple metrics.




\section{Prune rules}
The condition of a rule $\{C \Rightarrow T\}$ consists of variable-value pairs, $a_1 = v_1$, ..., $a_i = v_i$, ..., $a_K = v_K$, where $a_i$ and $v_i$ is the $i^{th}$ variable-value pair in the condition, and $K$ ($K\geq 1$) is the total number of variable-value pairs in the condition. A rule extracted from trees may include irrelevant variable-value pairs. This section provides methods to prune irrelevant variable-value pairs from rule conditions.

Let $E$ denote a metric measuring the quality of a rule, and a smaller value of $E$ indicates a better rule. Examples of $E$ include error of applying a rule to the training data, error of applying a rule to a validation data set, and pessimistic error that is a combination of training error and model complexity. Here we assume $E\geq 0$.


Let $E_0$ denote the $E$ of the original rule $\{C \Rightarrow T\}$, and let $E_{-i}$ denote the $E$ of the rule that leave the $i^{th}$ variable-value pair out. We use $decay_i$ to evaluate the effect of removing the $i^{th}$ pair:

 \begin{equation} \label{eq:decay1}
decay_i = \frac{E_{-i} - E_0}{max(E_0,s)}
\end{equation}


where $s$ is a positive number that bound the value of $decay_i$ when $E_0$ is 0 or very small. In the inTree R package, we currently set $s=10^{-6}$.

Equation \ref{eq:decay1} can be interpreted as the \textbf{relative} increase of error after removing a variable-value pair. Alternatively, one can define the decay function as the increase of error shown in Equation \ref{eq:decay2}.
 \begin{equation} \label{eq:decay2}
decay_i = E_{-i} - E_0
\end{equation}

When $decay_i$ is smaller than a threshold, e.g., 0.05, the $i^{th}$ variable-value pair may be considered unimportant for the rule and thus can be removed. We call this pruning method \textbf{leave-one-out pruning}.

Leave-one-out pruning is applied to each variable-value pair sequentially. Consider example: \{$X_1 = 0$ $\&$ $X_2 = 1$ $\&$ $X_{3} = 0$ $\Rightarrow$ \emph{true}\}. Use Equation \ref{eq:decay1} as the decay function, and set $s$ as 0.001 and the decay threshold as 0.05. Start with the last variable-value pair, assuming $E_0=0.1$, $E_{-3}=0.2$, then $decay_3$ = $\frac{0.2-0.1}{0.1}$ = $1$. As $decay_3$ $>$ 0.05, the last pair remains in the rule, and thus $E_0$ remains 0.1. Next consider the second pair $X_2 = 1$. Assuming $E_{-2}=0.104$, and thus $decay_2$ = $\frac{0.104-0.1}{0.1}$ = $0.04$ $<$ 0.05, then $X_2 = 1$ is removed from the rule. Now the rule becomes \{$X_1 = 0$ $\&$ $X_{3} = 0$ $\Rightarrow$ \emph{true}\}, and $E_0=0.104$. Finally consider the first pair $X_1 = 0$ in \{$X_1 = 0$ $\&$ $X_{3} = 0$ $\Rightarrow$ \emph{true}\}. Assuming $E_{-1}=0.3$, then $decay_1$ = $\frac{0.3-0.104}{0.104}$ $>$ 0.05, and thus the first pair remains in the rule. And the rule after pruning is \{$X_1 = 0$ $\&$ $X_{3} = 0$ $\Rightarrow$ \emph{true}\}. In this work we don't optimize the order of the pairs to be considered for pruning.



\section{Select Rules}
\subsection{Rule selection via feature selection}
The number of rules extracted from a tree ensemble can be large. We can rank the rules by error and frequency, however, the top rules could be similar to each other, i.e., redundant. It would be desirable to derive a compact rule set that contains relevant and non-redundant rules. \cite{deng2014cbc} uses feature selection to select a set of relevant and non-redundant conditions from associative classification rules \cite{liubing1998}. Here we apply feature selection to the conditions from a tree ensemble. The idea is creating a new data set $I$ by the following way described in \cite{deng2014cbc}.

Let $\{c_1, c_2,..., c_J\}$ denote the condition set of a rule set. Let $I_{ij}$ denote whether a condition $c_j$ is satisfied by the $i^{th}$ instance, that is,
\begin{equation}
I_{ij} = \begin{cases}
 1 & \text{$c_j$ is satisfied for the $i^{th}$ instance} \\
 0 & \text{otherwise} \\
\end{cases}
\end{equation}
A new data set $I$ is formed by combining the binary variables and the target: $\{[I_{i1},...,I_{iJ},t_i],i=1,...n\}$ where $t_i$ is the target value of the $i^{th}$ instance, and $n$ is the number of instances for training. Let $\{I_1, I_2,..., I_J\}$ denote the predictor variables. Then feature selection can be applied to the new data set to select a set of relevant and non-redundant variables, each essentially presenting a condition. More details of the algorithm can be found in \cite{deng2014cbc}. Furthermore, once conditions is selected, one can use the approach introduced in Section \ref{sec:extract} to assign outcomes to the conditions.


\subsection{Complexity-guided condition selection}
In \cite{deng2014cbc} the condition complexity (length of a condition) was not considered in the feature selection process. Here we consider condition complexity in the feature selection process by using the guided regularized random forest (GRRF) algorithm \cite{deng2013gene}.

In a regularized random forest (RRF), the regularized information gain in a tree node is defined as

\begin{equation}\label{eq:gain}
Gain_R(X_i) =
\begin{cases}
 \lambda_i \cdot Gain(X_i) & \text{$X_i \notin F$} \\
 Gain(X_i) & \text{$X_i \in F$} \\
\end{cases}
\end{equation}

where $F$ is the set of indices of variables used for splitting the previous nodes and is an empty set at the root node in the first tree. $Gain(\cdot)$ represents an ordinary information gain metric in decision trees. $\lambda_i \in(0,1]$ is called the penalty coefficient. The regularized information gain penalizes the $i^{th}$ variable when $i \notin F$, that is, the $i^{th}$ variable was not used in previous nodes. The regularized random forest adds the index of a new variable to $F$ if the variable adds enough new predictive information to the already selected variables.

A smaller $\lambda_i$ leads to a larger penalty. In RRF, $\lambda_i$ is the same for all variables. GRRF \cite{deng2013gene} assigns a value to $\lambda_i$ based on the global importance score of $X_i$ calculated from an ordinary random forest (RF). A larger importance score of $X_i$ would lead to a larger $\lambda_i$. Therefore, given two variables that are not selected in previous nodes and have similar information gain, the variable with a larger global importance score has a larger regularized information gain.

Here we further extend the concept of GRRF to handle rule complexity.
Particularly, $\lambda_i \in(0,1]$ is calculated based on the length of the condition represented by $i^{th}$ variable. That is,
\begin{equation}\label{eq:lambda}
 \lambda_i = \lambda_0 \cdot (1 - \gamma\ \cdot \frac{l_i}{l^*})\\
\end{equation}
where $\lambda_0\in(0,1]$ is called the base coefficient, $l_i$ is the length of the condition represented by the $i^{th}$ variable in $I$, $l^*$ is the maximum length of the conditions in the condition set, and $\gamma\in[0,1]$ controls the weight of the regularization on condition length.  It can be seen a longer rule condition leads to a smaller $\lambda_i$, and thus is penalized more in Equation \ref{eq:gain}. In addition, it may be also useful to add the global importance score to Equation \ref{eq:lambda} as follows

\begin{equation}\label{eq:comb}
 \lambda_i = \lambda_0 \cdot (1 - \gamma\ \cdot \frac{l_i}{l^*} + \beta \cdot imp_i)\\
\end{equation}
where $imp_i$ is the normalized importance score ($0\leq imp_i\leq1$) of $X_i$, calculated from an ordinary RF, and $\beta \in[0,1]$ controls the weight of the global importance scores.

Also, GRRF not only can select a subset of variables, but can also provide a score for each variable. However, the score can be biased in favour of variables entering the variable subset $F$ earlier in the feature selection process. One can build an ordinary RF on the selected variables, to calculate scores for the selected conditions.

\section{Variable interaction extraction based on association rule analysis}
Here we introduce methods to extract frequent variable interactions in a tree ensemble using association rule analysis \cite{agrawal1994fast}.

Association rule analysis \cite{agrawal1994fast,hahsler2007introduction} discovers associations between items in transaction data sets, where one transaction consists of one or multiple items. An example of an association rule is \{$bread$, $butter$\ $\Rightarrow$ $milk$\}, that is, transactions contain bread and butter also contain milk. The left-hand side of a rule is called the condition, and the right-hand side is called the outcome. An association rule can be measured by the following three metrics. \textbf{Support} is defined as the proportion of transactions containing the condition. \textbf{Confidence} is defined as the number of transactions containing both the condition and the outcome divided by the number of transactions containing the condition. \textbf{Length} is defined as the number of items contained in an association rule. For example, \{$bread$, $butter$\ $\Rightarrow$ $milk$\} has a length of 3.

Each rule from tree ensemble has the form of $C$ $\Rightarrow$ $T_k$ consists of variable-value pairs and the target-value pair, i.e., $a_1 = v_1$, ..., $a_i = v_i$, ..., $a_K = v_K$ and $T=T_k$. One can treat each variable-value pair $a_i = v_i$ or target-value pair $T=T_k$ as an item. Association rule analysis can then be used to obtain a set of association rules with a minimum support and confidence, but with a restriction that the left-hand side of the association rule contains only variable-value pairs, and the right-hand side contains only the target-value pair. Now one can rank the association rules by any one of metrics: support, confidence, length, or a combination of multiple metrics. Note the variable-value pairs contained in the association rule(s) with the largest support and length $l$ ($l\geq2$) can be considered as the most frequent $l$-way variable interactions.

The concept of ``item" in association rule can be used flexibly. For example, the variable of an variable-value pair can also be an item. This can be useful for the following situation. Numeric variables can have many combinations of variable-value pairs in the rule sets extracted from a tree ensemble. If we consider variable-values or target-value pairs as items, the support of association rules containing the variable-value pair of the numeric variables can be relatively small. To solve this issue, one can discretize the numeric variables. Alternatively, one can treat numeric variables, and variable-value pairs for discrete variables as items.

\section{Simplified tree ensemble learner (STEL)}
The rules extracted from a tree ensemble can be summarized into a rule-based learner, referred to as a simplified tree ensemble learner (STEL). \cite{friedman2008predictive} used a linear model summarizing the rules from a tree ensemble. Also, ideas in summarize associative classification rules into classifiers \cite{liubing1998,deng2014cbc} could be applied to tree ensemble rules. Here we introduce one algorithm to summarize rules.

Let $\Re$ denote an ordered rule list, and $\Re=\{\}$ at the beginning of the algorithm. The goal is to build a list of rule in $\Re$ ordered by priority. Then the rules in $\Re$ are applied, from the top to the bottom, to a new instance until a rule condition is satisfied by the instance. The right-hand side (i.e., outcome) of the rule then becomes the prediction for the new instance.

First define the default rule as $\{\} $ $\Rightarrow$ $t^*$, where $t^*$ is the most frequent class in the training data for classification, and the mean for regression. The default rule assigns $t^*$ to an instance's outcome disregarding the predictor variable values of the instance. 

Let $S$ denote the rule set including the two default rules and rules extracted from a tree ensemble. Rules below a certain frequency, e.g., 0.01, are removed from $S$ to avoid overfitting. Let $D$ denote the set including all the training instances at the start of the algorithm.  

The algorithm consists of multiple iterations. At each iteration, the best rule in $S$ evaluated by data $D$ is selected and added to the end of $\Re$. The best rule is defined as the rule with the minimum error. If there are ties, the rule with higher frequency is preferred. Preference on rules with shorter conditions is used to break further ties.
Then the data instances satisfying the best rule are removed from $D$, and the default rule and the rule metrics are updated based on the data instances left. If the default rule is selected as the best rule in an iteration, the algorithm returns $\Re$. If no instance is left, the default rule is assigned with the most frequent class for classification or mean for regression in the original training data. The default rule is added to the end of $\Re$, and the algorithm returns $\Re$.

Algorithms in summarizing associative classification rules may be useful in summarizing the rules from tree ensembles. For example, one can apply a decision tree algorithm to the new data set $I$, which summarizes the conditions into a rule-based classifier \cite{deng2014cbc}. But note the rules from tree ensembles are different than associative classification rules in that a variable or target in a tree ensemble rule can be numeric or discrete, while associative classification rules only handle discrete variables and target.

\section{Transform regression rules to classification rules}
A regression rule with an numeric outcome, e.g.,  \{$X_1=1 \Rightarrow 100$\}, may not be as interpretable as classification rules, e.g., \{$X_1=1 \Rightarrow large$\}. In the inTrees framework, one can easily transform regression rules to classification rules as follows. Firstly, build a tree ensemble for a regression problem, and extract rule conditions from the tree ensemble. Secondly, replace the numeric target variable with a discretized version (e.g., discretization with equal frequency). Then the methods dealing with classification rules can be used for the conditions extracted. Therefore, one can transform the regression rules to classification rules without re-building a tree ensemble after discretizing the target variable.

\section{Illustrative Examples}

Here we illustrate the inTrees functions by examples. Consider the following team optimization problem. One chooses a team with 10 players from 20 to play a game against another team. The team would win if a) either player 1 or player 2 is in the team; and b) Player 1 and player 2 do not play together in the team. Otherwise, the team would lose. The logic is shown in table \ref{table:twoperson}.

\begin{table}[h]
\centering
\caption{The team optimization example where the team would win only if either player 1 or player 2 plays, and they do not play together. In the table ``Y" stands for playing in the team, and ``N" stands for not playing.   \label{table:twoperson}}
\begin{tabular}{ccccc}
  \hline
player 1 & player 2 & player 3 & ... & outcome \\
  \hline
N & N & ... & ... & lose \\
  Y & Y & ... & ... & lose \\
  Y & N & ... &... & win \\
  N & Y & ... & ... & win \\
   \hline
\end{tabular}
\end{table}

We simulated the data set with 20 predictive variables $X_1$, ..., $X_{20}$, the target $T$, and 100 instances ($X_i$ represents player $i$). For each instance, 10 variables are randomly selected and assigned value ``Y", and other variables are assigned value ``N". $T$ = ``win" when one and only one of $X_1$ and $X_2$ equals ``Y", and $T$ = ``lose" otherwise.

In the following we apply the inTrees framework to the data set. We build a regularized random forest (RRF) with 100 trees on the data set. 1923 rule conditions with maximum length of 6 were extracted from the RRF by the condition extraction method in the inTrees package. After de-duping the same rules, there are 1835 unique rule conditions.

Next, inTrees provides functions to assign outcomes to the conditions, and, in the mean time, calculates the length, frequency and error of each rule. Alternatively, one can extract rules from a random forest using algorithm \ref{algo:extractrule}. However, the original outcome assignments of the rules from RF may not be reliable because each rule is evaluated by a sub-sample of training data.

Table \ref{tab:rule} shows two examples of rules consisting conditions and predictions (outcome assignments), and the rule metrics: length, frequency, and error. In the table ``X[,$k$] \%in\% c($v$)" represents that the $k^{th}$ variable equals one of the values from set $v$ when $X[,k]$ is categorical. When a variable is numeric, ``$\leq$" or ``$\geq$" is used as the relation operator between the variable and a numeric value. The conditions are presented in an R language format, and thus can be directly executed in R given ``X" is the predictor variable matrix (return ``TRUE" for instances satisfying the conditions, and ``FALSE" for other instances).

\begin{table}[h]
\scriptsize
\caption{Two examples of rules consisting of conditions and predictions (denoted by ``pred"), and the metrics of rules. \label{tab:rule}}
\centering
\begin{tabular}{lllll}
  \hline
len & freq & err & condition & pred \\
  \hline
3 & 0.07 & 0 & X[,1] \%in\% c('N') \& X[,2] \%in\% c('N') \& X[,19] \%in\% c('N') & lose \\
  3 & 0.16 & 0 & X[,1] \%in\% c('Y') \& X[,2] \%in\% c('N') \& X[,19] \%in\% c('N') & win \\
   \hline
\end{tabular}
\end{table}

Next discuss pruning each rule. The rules shown in Table \ref{tab:rule} after being pruned have shorter conditions shown in Table \ref{tab:ruleprune}. With the irrelevant variable-value pairs being removed from the conditions, the frequency of the both rules increases without increase of error.

\begin{table}[h]
\scriptsize
\caption{Pruned rules for the rules shown in Table \ref{tab:rule}. \label{tab:ruleprune}}
\centering
\begin{tabular}{lllll}
  \hline
len & freq & err & condition & pred \\
  \hline
2 & 0.22 & 0 & X[,1] \%in\% c('N') \& X[,2] \%in\% c('N') & lose \\
  2 & 0.24 & 0 & X[,1] \%in\% c('Y') \& X[,2] \%in\% c('N') & win \\
   \hline
\end{tabular}
\end{table}

Now consider rule selection. We applied the complexity-guided regularized random forest to the rule set (with each rule being pruned). Table \ref{tab:ruleselect} shows the selected rules and their metrics.

\begin{table}[h]
\scriptsize
\caption{Rule set selected by guided regularized random forest. \label{tab:ruleselect}}
\centering
\begin{tabular}{lllll}
  \hline
len & freq & err & condition & pred  \\
  \hline
2 & 0.22 & 0 & X[,1] \%in\% c('N') \& X[,2] \%in\% c('N') & lose \\
  2 & 0.22 & 0 & X[,1] \%in\% c('Y') \& X[,2] \%in\% c('Y') & lose  \\
  2 & 0.24 & 0 & X[,1] \%in\% c('Y') \& X[,2] \%in\% c('N') & win  \\
  2 & 0.32 & 0 & X[,1] \%in\% c('N') \& X[,2] \%in\% c('Y') & win  \\
   \hline
\end{tabular}
\end{table}

Furthermore, one can build an ordered list of rules that can be used as a classifier, i.e., the simplified tree ensemble learner (STEL). Table \ref{eq:rulecls} shows the classifier. A more readable version of the classifier is presented in Table \ref{tab:presentrulecls}.

\begin{table}[h]
\scriptsize
\caption{An ordered list of rules as a classifier. Note the frequency (denoted by ``freq") and error of (denoted by ``err") of
 a rule in the classifier are calculated at the iteration the rule is selected, and so can be different than applying the rules individually to the whole training data. These conditions are R-executable given X is the data including the predictive variables. The last rule ``X[,1]==X[,1]" with prediction of TRUE simply means ``All instances that do not satisfy the above rules are predicted as lose". \label{eq:rulecls}}
\centering
\begin{tabular}{lllll}
  \hline
len & freq & err & condition & pred \\
  \hline
2 & 0.32 & 0 & X[,1] \%in\% c('N') \& X[,2] \%in\% c('Y') & win \\
  2 & 0.24 & 0 & X[,1] \%in\% c('Y') \& X[,2] \%in\% c('N') & win \\
  1 & 0.44 & 0 & X[,1]==X[,1] & lose \\
   \hline
\end{tabular}
\end{table}

\begin{table}[h]
\scriptsize
\caption{Present STEL rules in a more readable form. \label{tab:presentrulecls}}
\centering
\begin{tabular}{lllll}
  \hline
len & freq & err & condition & pred \\
  \hline
2 & 0.32 & 0 & X1 \%in\% c('N') \& X2 \%in\% c('Y') & win \\
  2 & 0.24 & 0 & X1 \%in\% c('Y') \& X2 \%in\% c('N') & win \\
  1 & 0.44 & 0 & Else & lose \\
   \hline
\end{tabular}
\end{table}



One can also use inTrees to extract frequent variable interactions from the rules extracted from a tree ensemble (without removing identical rules, pruning rules or selecting rules). The top 10 most frequent combinations of variable-value pairs (length $\geq$ 2) are shown in the ``condition" column of table \ref{eq:rulefreq}, as a combination of variable-value pairs is essentially a condition. It can be seen that the top 4 conditions capture the true patterns. About 4\% (support) of the rules from the random forest contain each of the top 4 conditions, and 100\% of the rules containing one of the top 4 conditions have the true outcomes (confidence = 1). The fifth and sixth conditions in the table have support more than 3\% but confidence less than 70\%. Therefore these frequent variable interactions do not lead to consistent outcome assignments. This indicates one should look at both the frequency and the confidence when considering variable interactions. 

\begin{table}[h]
\scriptsize
\caption{Top 10 most frequent combinations of variable-value pairs with length $\geq$ 2 appearing in the tree ensemble, and their metrics. Column ``len" measures the length of the conditions (or variable interactions)\label{eq:rulefreq}}
\centering
\begin{tabular}{lllll}
  \hline
len & sup & conf & condition & pred \\
  \hline
2 & 0.046 & 1 & X[,1] \%in\% c('N') \& X[,2] \%in\% c('N') & lose \\
  2 & 0.044 & 1 & X[,1] \%in\% c('Y') \& X[,2] \%in\% c('N') & win \\
  2 & 0.041 & 1 & X[,1] \%in\% c('N') \& X[,2] \%in\% c('Y') & win \\
  2 & 0.039 & 1 & X[,1] \%in\% c('Y') \& X[,2] \%in\% c('Y') & lose \\
  2 & 0.034 & 0.699 & X[,12] \%in\% c('N') \& X[,5] \%in\% c('Y') & win \\
  2 & 0.032 & 0.667 & X[,12] \%in\% c('Y') \& X[,19] \%in\% c('Y') & lose \\
  2 & 0.029 & 0.696 & X[,11] \%in\% c('Y') \& X[,19] \%in\% c('Y') & lose \\
  2 & 0.028 & 0.635 & X[,11] \%in\% c('Y') \& X[,12] \%in\% c('Y') & lose \\
  2 & 0.026 & 0.595 & X[,12] \%in\% c('N') \& X[,5] \%in\% c('N') & lose \\
  2 & 0.025 & 0.615 & X[,19] \%in\% c('Y') \& X[,9] \%in\% c('Y') & lose \\
   \hline

\end{tabular}
\end{table}

\begin{table}[h]
\scriptsize
\caption{Summary of 20 data sets, and error rates of rpart and the simplified tree ensemble learner (STEL) for each data set. Relative differences are calculated.  The data sets with a significant difference at the 0.05 level are marked with ``$\circ$" when STEL has higher error rates than rpart or ``$\bullet$" when STEL has lower error rates. STEL outperforms rpart for 13 data sets but only loses for 5 data sets (with a statistically significant difference). Furthermore, the relative differences tend to be larger when STEL has a advantage. \label{tab:err}}
\centering
\begin{tabular}{|r|cc|cl|r|}
  \hline
 & numInst & numFea & rpart & STEL & difference(\%) \\
  \hline
anneal & 898 & 38 & 0.098 & 0.070 $\bullet$  & 28.7\% \\
  austra & 690 & 14 & 0.145 & 0.157 $\circ$ &  8.0\% \\
  auto & 205 & 25 & 0.376 & 0.262 $\bullet$  & 30.2\% \\
  breast & 699 & 10 & 0.058 & 0.048 $\bullet$  & 17.4\% \\
  crx & 690 & 15 & 0.148 & 0.159 $\circ$ &  7.2\% \\
  german & 1000 & 20 & 0.274 & 0.286 $\circ$ &  4.3\% \\
  glass & 214 & 9 & 0.342 & 0.310 $\bullet$  &  9.2\% \\
  heart & 270 & 13 & 0.219 & 0.224 &  2.1\% \\
  hepati & 155 & 19 & 0.209 & 0.211 &  0.6\% \\
  horse & 368 & 22 & 0.164 & 0.197 $\circ$ & 16.6\% \\
  iris & 150 & 4 & 0.064 & 0.047 $\bullet$  & 26.6\% \\
  labor & 57 & 16 & 0.223 & 0.148 $\bullet$  & 33.7\% \\
  led7 & 3200 & 7 & 0.318 & 0.269 $\bullet$  & 15.3\% \\
  lymph & 148 & 18 & 0.268 & 0.209 $\bullet$  & 21.9\% \\
  pima & 768 & 8 & 0.260 & 0.272 $\circ$ &  4.4\% \\
  tic-tac & 958 & 9 & 0.094 & 0.002 $\bullet$  & 97.9\% \\
  vehicle & 846 & 18 & 0.325 & 0.285 $\bullet$  & 12.2\% \\
  waveform & 5000 & 21 & 0.262 & 0.198 $\bullet$  & 24.2\% \\
  wine & 178 & 13 & 0.122 & 0.086 $\bullet$  & 29.8\% \\
  zoo & 101 & 16 & 0.211 & 0.061 $\bullet$  & 71.3\% \\
   \hline
\end{tabular}
\end{table}

\section{Experiments}

To test the accuracy of the simplified tree ensemble learner (STEL) implemented in the ``inTrees" package version 1.0, we compare it to the popularly used decision tree method in the rpart package \cite{breiman1984,therneau2010rpart} on 20 data sets from UCI\cite{blake1998uci}. Random forest with 100 trees was used. Conditions with maximum length of 6 were extracted from random forest. Also, for data sets with more than 2000 conditions extracted, 2000 conditions were randomly sampled and used. For each data set, we randomly sampled 2/3 of the data for training the models, and used the rest 1/3 for testing the models. The procedure was performed 100 times (each run is different due to randomness). The average error rate for each method and the relative difference between the two methods (defined as the difference between the larger error and lower error divided by the larger error) on each data set were calculated. The paired t-test was also applied to the error rates of the two methods from the 100 runs for each data set. The data sets with a significant difference between the two methods at the 0.05 level are marked with ``$\circ$" (STEL worse than rpart) or ``$\bullet$" (STEL better than rpart) in the table.

The results are shown in table \ref{tab:err}. STEL outperforms rpart with a statistically significant difference for 13 data sets, and loses with a statistically significant difference for only 5 data sets. Furthermore, when STEL outperforms rpart, most of the relative differences are greater than 10\%, while when STEL loses, only one data has a relative difference more than 10\% (16.6\%). 

Table \ref{tab:ucirule} shows the most accurate rule that has a frequency greater than 0.1 for each data set. Most rules have error rate of 0. The rule for ``led7" has error rate greater than 0.2 (i.e., 20\%), which may be because the data set is difficult to classify (Table \ref{tab:err} shows it indeed has relatively high error rate). 

\begin{table}[h]
\scriptsize
\centering
\caption{The most accurate rule with a minimum frequency of 0.1 for each UCI data set. ``bwnfp" for ``glass" data set stands for class ``building\_windows\_non\_float\_processed".}
\label{tab:ucirule}
\begin{tabular}{rlllll}
  \hline
 & len & freq & err & condition & pred \\
  \hline
anneal & 5 & 0.342 & 0 & X4$<$=1.5 \& X5$<$=82.5 \& X7 \%in\% c('S')  & 3 \\
       &   &       &   & \& X8$<$=2.5 \& X33$<$=0.7995 &  \\
  \hline
  austra & 5 & 0.181 & 0 & X5$<$=7.5 \& X7$<$=3.375 \& X8$<$=0.5  & 0 \\
       &   &       &   & \& X13$<$=415.5 \& X14$<$=251 &  \\
\hline
  auto & 5 & 0.195 & 0 & X1$>$71.5 \& X2 \%in\% c('bmw','honda','isuzu', & 0 \\
       &   &       &   & 'jaguar','mazda','nissan','peugot','subaru','toyota') &  \\
       &   &       &   & \& X5 \%in\% c('four') \& X10$<$=187.25 \& X21$>$69.5 &  \\
\hline
  breast & 3 & 0.591 & 0 & X3$<$=3.5 \& X7$<$=2.5 \& X9$<$=3.5 &  benign \\
\hline
  crx & 4 & 0.188 & 0 & X3$>$1.5625 \& X6 \%in\% c('aa','c','d','ff','i','j',  & no \\
      &   &       &   & 'k','m','r') \& X9 \%in\% c('f') \& X15$<$=492 &  \\
\hline
  german & 5 & 0.132 & 0.015 & X1 \%in\% c('no-account') \& X5$<$=4103.5  &  good \\
         &   &       &   & \& X10 \%in\% c(' guarantor',' none')  &  \\
         &   &       &   & \& X13$>$33.5 \& X14 \%in\% c(' none') &  \\
\hline
  glass & 5 & 0.136 & 0 & X1$<$=1.517325 \& X3$>$2.7 \& X4$>$1.42  & bwnfp\\
        &   &       &   & \& X7$>$7.82 \& X9$<$=0.16 &  \\
\hline
  heart & 4 & 0.2 & 0 & X1$<$=55.5 \& X4$>$119 \& X10$<$=1.7 \& X13$<$=4.5 & 1 \\
\hline
  hepati & 6 & 0.542 & 0 & X1$<$=61.5 \& X11$>$1.5 \& X13$>$1.5  & 2 \\
         &   &       &   & \& X14$<$=3.7 \& X15$<$=218.5 \& X18$>$40.5 &  \\
\hline
  horse & 5 & 0.188 & 0 & X3$<$=38.45 \& X3$>$37.25 \& X4$<$=126 & 1 \\
        &   &       &   &  \& X10$>$2.5 \& X12$>$2.5 &  \\
\hline
  iris & 1 & 0.333 & 0 & X3$<$=2.55 & setosa \\
\hline
  labor & 4 & 0.614 & 0 & X2$>$2.75 \& X7 \%in\% c('empl\_contr','ret\_allw')  & good \\
        &   &       &   & \& X8$>$5 \& X13 \%in\% c('yes') &  \\
\hline
  led7 & 3 & 0.103 & 0.211 & X1$<$=0.5 \& X2$<$=0.5 \& X6$>$0.5 & 1 \\
\hline
  lymph & 4 & 0.284 & 0 & X2$>$1.5 \& X13$>$2.5 \& X13$<$=3.5 \& X18$<$=2.5 & 2 \\
\hline
  pima & 3 & 0.124 & 0 & X2$<$=106.5 \& X6$<$=29.95 \& X8$<$=28.5 & 0 \\
\hline
  tic-tac & 3 & 0.225 & 0 & X1 \%in\% c('b','x') \& X5 \%in\% c('b','x')  & positive \\
          &   &       &   &  \& X9 \%in\% c('b','x') &  \\
\hline
  vehicle & 5 & 0.102 & 0 & X3$>$71.5 \& X6$>$8.5 \& X7$>$142.5  & 4 \\
          &   &       &   & \& X12$<$=376.5 \& X14$>$63.5 &  \\
\hline
  waveform & 5 & 0.102 & 0.059 & X6$<$=1.655 \& X9$<$=2.99  & 2 \\
           &   &       &   & \& X11$>$3.415 \& X12$>$2.49 \& X14$>$2.075 &  \\
\hline
  wine & 3 & 0.337 & 0 & X1$<$=12.78 \& X2$<$=4.575 \& X10$<$=4.84 & 2 \\
\hline
  zoo & 1 & 0.406 & 0 & X4$>$0.5 & 1 \\
   \hline
\end{tabular}
\end{table}

\section{Conclusions}
In this work we propose the inTree framework that consists of algorithms extracting, measuring, pruning, and selecting conditions/rules, and extracting frequent variable interactions/conditions from tree ensembles. Note the methods here can be applied to both classification and regression problems. The inTrees framework has been implemented in the ``inTrees" R package for tree ensembles: random forests, regularized random forests, and the generalized boosted regression models included in the ``gbm" R package \cite{ridgeway2006generalized}.

This work also demonstrates that the rules from tree ensembles can be accurate after being processed. This conclusion can be valuable to the rule mining area. For example, in the associative classification rule mining area (leveraging association rule \cite{agrawal1994fast} for classification), numerous algorithms have been proposed to select a subset of associative classification rules or summarize rules into a classifier \cite{liubing1998,deng2014cbc}. However, the predictor variables and the target need to be discrete (thus only for classification problems), and extracting associative classification rules can be computationally expensive. On the other hand, tree ensembles can be built and the rules can be extracted efficiently (also, both tree building and rule extraction can be performed in a distributed manner).  Furthermore, tree ensembles can be built on mixed categorical and numeric predictors, and can be used for both classification and regression problems. Note the ideas selecting and summarizing associative classification rules could still be applied to the rules extracted from tree ensembles.

\bibliographystyle{model1b-num-names}
\bibliography{ref}

\begin{thebibliography}{14}
\expandafter\ifx\csname natexlab\endcsname\relax\def\natexlab#1{#1}\fi
\providecommand{\url}[1]{\texttt{#1}}
\providecommand{\href}[2]{#2}
\providecommand{\path}[1]{#1}
\providecommand{\DOIprefix}{doi:}
\providecommand{\ArXivprefix}{arXiv:}
\providecommand{\URLprefix}{URL: }
\providecommand{\Pubmedprefix}{pmid:}
\providecommand{\doi}[1]{\href{http://dx.doi.org/#1}{\path{#1}}}
\providecommand{\Pubmed}[1]{\href{pmid:#1}{\path{#1}}}
\providecommand{\bibinfo}[2]{#2}
\ifx\xfnm\relax \def\xfnm[#1]{\unskip,\space#1}\fi
\bibitem[{Agrawal et~al.(1994)Agrawal, Srikant et~al.}]{agrawal1994fast}
\bibinfo{author}{R.~Agrawal}, \bibinfo{author}{R.~Srikant}, et~al.,
  \bibinfo{title}{Fast algorithms for mining association rules}, in:
  \bibinfo{booktitle}{Proc. 20th int. conf. very large data bases, VLDB},
  volume \bibinfo{volume}{1215}, pp. \bibinfo{pages}{487--499}.
\bibitem[{Blake and Merz(1998)}]{blake1998uci}
\bibinfo{author}{C.~Blake}, \bibinfo{author}{C.~Merz}, \bibinfo{title}{{UCI
  repository of machine learning databases}}  (\bibinfo{year}{1998}).
\bibitem[{Breiman(2001)}]{breiman2001random}
\bibinfo{author}{L.~Breiman}, \bibinfo{title}{Random forests},
  \bibinfo{journal}{Machine learning} \bibinfo{volume}{45}
  (\bibinfo{year}{2001}) \bibinfo{pages}{5--32}.
\bibitem[{Breiman et~al.(1984)Breiman, Friedman, Olshen and
  Stone.}]{breiman1984}
\bibinfo{author}{L.~Breiman}, \bibinfo{author}{J.~Friedman},
  \bibinfo{author}{R.~Olshen}, \bibinfo{author}{C.~Stone.},
  \bibinfo{title}{Classification and Regression Trees},
  \bibinfo{publisher}{Wadsworth, Belmont, MA}, \bibinfo{year}{1984}.
\bibitem[{Deng and Runger(2013)}]{deng2013gene}
\bibinfo{author}{H.~Deng}, \bibinfo{author}{G.~Runger}, \bibinfo{title}{Gene
  selection with guided regularized random forest}, \bibinfo{journal}{Pattern
  Recognition} \bibinfo{volume}{46} (\bibinfo{year}{2013})
  \bibinfo{pages}{3483--3489}.
\bibitem[{Deng et~al.(2014)Deng, Runger, Tuv and Bannister}]{deng2014cbc}
\bibinfo{author}{H.~Deng}, \bibinfo{author}{G.~Runger},
  \bibinfo{author}{E.~Tuv}, \bibinfo{author}{W.~Bannister},
  \bibinfo{title}{Cbc: An associative classifier with a small number of rules},
  \bibinfo{journal}{Decision Support Systems} \bibinfo{volume}{59}
  (\bibinfo{year}{2014}) \bibinfo{pages}{163--170}.
\bibitem[{Friedman(2001)}]{friedman2001greedy}
\bibinfo{author}{J.H. Friedman}, \bibinfo{title}{Greedy function approximation:
  a gradient boosting machine}, \bibinfo{journal}{Annals of Statistics}
  (\bibinfo{year}{2001}) \bibinfo{pages}{1189--1232}.
\bibitem[{Friedman and Popescu(2008)}]{friedman2008predictive}
\bibinfo{author}{J.H. Friedman}, \bibinfo{author}{B.E. Popescu},
  \bibinfo{title}{Predictive learning via rule ensembles},
  \bibinfo{journal}{The Annals of Applied Statistics}  (\bibinfo{year}{2008})
  \bibinfo{pages}{916--954}.
\bibitem[{Hahsler et~al.(2007)Hahsler, Gr{\"u}n and
  Hornik}]{hahsler2007introduction}
\bibinfo{author}{M.~Hahsler}, \bibinfo{author}{B.~Gr{\"u}n},
  \bibinfo{author}{K.~Hornik}, \bibinfo{title}{Introduction to arules--mining
  association rules and frequent item sets}, \bibinfo{journal}{SIGKDD
  EXPLORATIONS}  (\bibinfo{year}{2007}).
\bibitem[{Kaufman et~al.(2012)Kaufman, Rosset, Perlich and
  Stitelman}]{kaufman2012leakage}
\bibinfo{author}{S.~Kaufman}, \bibinfo{author}{S.~Rosset},
  \bibinfo{author}{C.~Perlich}, \bibinfo{author}{O.~Stitelman},
  \bibinfo{title}{Leakage in data mining: Formulation, detection, and
  avoidance}, \bibinfo{journal}{ACM Transactions on Knowledge Discovery from
  Data (TKDD)} \bibinfo{volume}{6} (\bibinfo{year}{2012}) \bibinfo{pages}{15}.
\bibitem[{Liaw and Wiener(2002)}]{liaw2002classification}
\bibinfo{author}{A.~Liaw}, \bibinfo{author}{M.~Wiener},
  \bibinfo{title}{Classification and regression by randomforest},
  \bibinfo{journal}{R news} \bibinfo{volume}{2} (\bibinfo{year}{2002})
  \bibinfo{pages}{18--22}.
\bibitem[{Liu et~al.(1998)Liu, Hsu and Ma}]{liubing1998}
\bibinfo{author}{B.~Liu}, \bibinfo{author}{W.~Hsu}, \bibinfo{author}{Y.~Ma},
  \bibinfo{title}{Integrating classification and association rule mining}, in:
  \bibinfo{booktitle}{Proceeding of the 1998 International Conference on
  Knowledge Discovery and Data Mining}, \bibinfo{publisher}{ACM},
  \bibinfo{year}{1998}, pp. \bibinfo{pages}{80--86}.
\bibitem[{Ridgeway(2006)}]{ridgeway2006generalized}
\bibinfo{author}{G.~Ridgeway}, \bibinfo{title}{Generalized boosted regression
  models}, \bibinfo{journal}{Documentation on the R Package ‘gbm’, version
  1{\textperiodcentered} 5} \bibinfo{volume}{7} (\bibinfo{year}{2006}).
\bibitem[{Therneau et~al.(2010)Therneau, Atkinson and
  Ripley}]{therneau2010rpart}
\bibinfo{author}{T.M. Therneau}, \bibinfo{author}{B.~Atkinson},
  \bibinfo{author}{B.~Ripley}, \bibinfo{title}{rpart: Recursive partitioning},
  \bibinfo{journal}{R package version} \bibinfo{volume}{3}
  (\bibinfo{year}{2010}).

\end{thebibliography}

\end{document}